\documentclass[letterpaper, 10 pt, conference]{ieeeconf}  

\IEEEoverridecommandlockouts                              

\usepackage{graphicx} 
\usepackage{subfigure}
\usepackage{amsmath} 
\usepackage{amssymb}  
\usepackage{multirow}
\usepackage[dvipsnames]{xcolor}
\definecolor{c_curb}{RGB}{196, 196, 196}
\definecolor{c_crosswalk}{RGB}{200, 140, 140}
\definecolor{c_road}{RGB}{128, 64, 128}
\definecolor{c_sidewalk}{RGB}{232, 35, 244}
\definecolor{c_building}{RGB}{70, 70, 70}
\definecolor{c_person}{RGB}{60, 20, 220}
\definecolor{c_bicyclist}{RGB}{0, 0, 255}
\definecolor{c_motorcyclist}{RGB}{100, 0, 255}
\definecolor{c_sky}{RGB}{180, 130, 70}
\definecolor{c_vegetation}{RGB}{35, 142, 107}
\definecolor{c_manhole}{RGB}{160, 128, 100}
\definecolor{c_pole}{RGB}{153, 153, 153}
\definecolor{c_traffic_sign}{RGB}{0, 220, 220}
\definecolor{c_bicycle}{RGB}{32, 11, 119}
\definecolor{c_bus}{RGB}{100, 60, 0}
\definecolor{c_car}{RGB}{142, 0, 0}
\definecolor{c_motorcycle}{RGB}{230, 0, 0}
\definecolor{c_truck}{RGB}{70, 0, 0}

\title{\LARGE \bf
Probabilistic Semantic Mapping for Urban Autonomous Driving Applications
}

\author{David Paz$^{*}$, Hengyuan Zhang$^{*}$, Qinru Li$^{*}$, Hao Xiang$^{*}$, Henrik I. Christensen
\thanks{$^{*}$These members contributed equally to this publication.}
\thanks{Affiliation: Contextual Robotics Institute,
        University of California, San Diego, 9500 Gilman Dr, La Jolla, CA 92093}
}

\begin{document}

\maketitle
\thispagestyle{empty}
\pagestyle{empty}

\begin{abstract}

Recent advancements in statistical learning and computational abilities have enabled autonomous vehicle technology to develop at a much faster rate. While many of the architectures previously introduced are capable of operating under highly dynamic environments, many of these are constrained to smaller-scale deployments, require constant maintenance due to the associated scalability cost with high-definition (HD) maps, and involve tedious manual labeling. As an attempt to tackle this problem, we propose to fuse image and pre-built point cloud map information to perform automatic and accurate labeling of static landmarks such as roads, sidewalks, crosswalks, and lanes. The method performs semantic segmentation on 2D images, associates the semantic labels with point cloud maps to accurately localize them in the world, and leverages the confusion matrix formulation to construct a probabilistic semantic map in bird's eye view from semantic point clouds. Experiments from data collected in an urban environment show that this model is able to predict most road features and can be extended for automatically incorporating road features into HD maps with potential future work directions.

\end{abstract}

\section{INTRODUCTION}


High-definition (HD) maps provide useful information for autonomous vehicles to understand the static parts of the scene. 
Due to the nature of the information encoded in HD maps--such as centimeter-level definitions for road networks, traffic signs, crosswalks, stop signs, traffic lights, and even speed limits--many of these maps become outdated during construction or road network changes.
Given these fast-changing environments, manually annotated HD maps become obsolete and may cause vehicles to perform inadequate reference path tracking actions leading to unsafe scenarios. In the process of HD map generation, extracting semantics and attributes from data takes the most amount of the work \cite{Jiao18MLHDMap}. A model that automates this process could improve HD map generation, reduce labor costs, and increase driving safety. 

Retrieval of centimeter-level semantic labels of the scene is a non-trivial task. Prior work such as \cite{Douillard11Classification, Sengupta12DenseVisual} adopted Conditional Random Fields (CRF) to assign semantic labels. The advancement of deep learning also provides promising results in terms of retrieving semantic information from images. State of the art semantic segmentation algorithms such as \cite{Long_2015_CVPR_FCNNet,Zhao_2017_PSPNet,Chen_2018_DeeplabV3Plus} generate pixel-level semantic labels with greater accuracy. Researchers have also explored methods to create semantic maps of the environment: examples are given in \cite{Maturana18RealtimeSemantic, mattyus2017deeproadmapper, homayounfar2019dagmapper}. Multi-sensor fusion has been used to improve the robustness of these algorithms; however, these approaches either use aerial imagery to extract road information or do not explicitly map the lane and crosswalk information, which are required for HD maps. Therefore, a detailed semantic map for urban autonomous vehicle applications is still of interest to explore.

To address these gaps, our work is focused on leveraging dense point maps built from a 16 channel LiDAR and state of the art semantically labeled images from deep neural networks, trained only on a publicly available dataset, to automatically generate dense probabilistic semantic maps in urban driving environments that provide robust labels for roads, lane marks, crosswalks, and sidewalks. We create a bird's eye view (BEV) semantic map of the environments by modeling the uncertainty of the semantic segmentation network with a confusion matrix formulation. Furthermore, the fusion of LiDAR intensity slightly improves the accuracy of lane mark segmentation on the road. The comparison with a ground truth HD map that has been tested in our autonomous vehicle for campus mail delivery tasks shows that the proposed model can identify semantic features on the road and localize them accurately. 

\begin{figure*}
    \centering
    \includegraphics[scale=0.56]{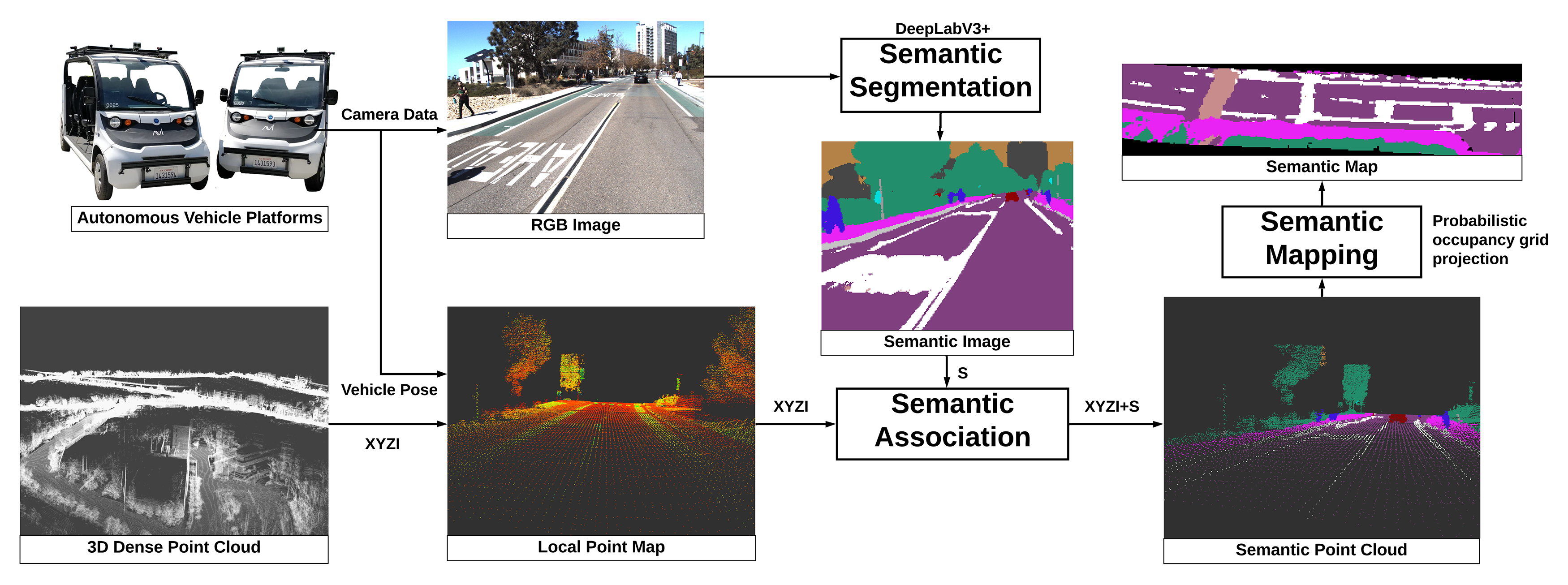}
    \caption{Our pipeline for generating probabilistic semantic maps for road feature extraction and HD mapping applications. Our model consists of three parts: 1) Semantic Segmentation: predict semantic labels based on 2D images. 2) Semantic Association: associate point clouds with predicted semantic labels. 3) Semantic Mapping: use a probabilistic semantic mapping method to capture the latent distribution of each label and utilize LiDAR intensity to augment lane mark prediction.}
    \label{fig:pipeline}
\end{figure*}

\section{RELATED WORK}

\textbf{Semantic Segmentation:}
Semantic segmentation is the task of assigning each observed data point (e.g. pixel or voxel) to a class label that contains semantic meanings. Research in this field has made tremendous progress as large scale datasets like CityScapes \cite{Cordts_2016_CVPR_cityscapes}, CamVid \cite{BrostowFC:PRL2008_Camvid}, Mapillary\cite{MVD2017_Mapillary_Vistas}, become available. Given that HD maps require fine-grained labeling for each object, building such maps can significantly benefit from the pixel level information provided by semantic segmentation algorithms. 

In 2D semantic segmentation, predominant works \cite{Long_2015_CVPR_FCNNet,Zhao_2017_PSPNet,chen2017rethinking_deeplabv3} leverage pyramid like encoder-decoder architecture to capture both global and local information in the images. Trained on the large scale datasets aforementioned, these network architectures can easily detect objects on the road even when these objects only have textural differences like colors. In 3D semantic segmentation, work has also been done by modeling 3D LiDAR point clouds as range images and then feeding them to a classical CNN network to classify each point\cite{Wu_2018_squeezeseg,heinzler2020cnn_rangenet,wang2018pointseg}. While the results of these works seem promising, due to the nature of LiDAR sensors, these methods cannot distinguish objects with texture differences such as colors. Researchers have also proposed an alternative approach of segmentation on voxelized point clouds \cite{Tchapmi_2017_segcloud}. However, such 3D convolutions are computationally expensive and usually require dense raw point cloud measurements (i.e 32 or 64-channel LiDARs), making it troublesome to operate in real time.


\textbf{Semantic Mapping:}
Semantic mapping has a rich meaning in various literature \cite{KOSTAVELIS201586}. We adopt the definition from \cite{Wolf08SemanticMapping}, which is the process of building maps that represent not only an occupancy metric but also other properties of the environment. In the context of autonomous driving, the drivable areas and road features are often included.

Other methods like \cite{Sengupta12DenseVisual} propose CRF based methods in dense semantic mapping. They use associative hierarchical CRF  for semantic segmentation and pairwise CRF for mapping. The pairwise potential minimization enforces the output smoothness. Sengupta et. al. \cite{Sengupta13Stereo} utilize a stereo pair to estimate the depth robustly, but they do not explicitly map the lane and crosswalk information, which are required for HD maps. 

Daniel et. al. \cite{Maturana18RealtimeSemantic} fuse semantic images from a camera with LiDAR point clouds, but they use real-time raw point clouds from a 64-channel LiDAR providing much denser real-time information whereas we can build map from a 16-channel LiDAR that has lower cost. Additionally, their focus is on off-road terrains, in contrast to our focus: the urban driving scenario. Urban driving scenes require special treatment of classes that contain specific traffic rules, such as lane marks and crosswalks.

\textbf{Probabilistic Map}: Probabilistic maps have been successfully applied to localization \cite{Levinson10ProbMap} \cite{Shao2017HIGHACCURACYVL} and pedestrian motion prediction \cite{Wu18ProbMapPedes}. Probabilistic maps can capture the inherent distribution information in a discrete space while filtering the noise. In this work, we successfully apply these techniques to semantic map generation while leveraging the prior information in LiDAR's intensity channel to produce more stable semantic maps.

\section{METHOD}

Our model consists of three parts: semantic segmentation, semantic association, and semantic mapping. The overall architecture is shown in Figure \ref{fig:pipeline}. We use semantic segmentation networks to predict semantic labels on 2D images and then associate semantic labels with densified 3D point clouds. Afterwards we apply probabilistic mapping to capture the distribution of labels assigned to each grid. In this section, we will describe each part in detail.

\subsection{Image Semantic Segmentation}

We use the DeepLabV3Plus \cite{Chen_2018_DeeplabV3Plus} network architecture to extract the semantic segmentation from 2D images. A lightweight ResNeXt50 \cite{Xie_2017_ResNeXt} pre-trained on ImageNet \cite{ImageNet2015} is used as our feature extraction backbone. Compared with other backbones like ResNet101 \cite{He_2016_resnet}, it can achieve the same mean of intersection over union (mIoU) value with fewer parameters and faster inference times. We also adopt the depth-wise separable convolution inspired by \cite{Chen_2018_DeeplabV3Plus, Chollet_2017_Xception} in our spatial pyramid layers and decoder layers to further improve the inference time while preserving the same performance.  

Our semantic segmentation network is trained in the Mapillary Vistas dataset \cite{MVD2017_Mapillary_Vistas}. This dataset provides a large number of pixel-level semantic segmented images with 66 different kinds of labels in autonomous vehicle scenarios. We reduce the labels into 19 classes by removing labels that are not essential in our driving environment (e.g. snow) and merging labels with similar semantic meanings together (e.g. zebra line and crosswalk). This decision is made based on the observation that some classes are unlikely to appear in our test environment. The details of label merging can be found in Section \ref{sec:exp_semantic}. All the training labels and their associated colors are presented in Table \ref{table:training_labels}.

\subsection{Point Cloud Semantic Association}

Given a semantic image, estimating the relative depth for the semantic pixel data can help us reconstruct the 3D scene with semantic labels. This information, however, is usually not available. Depth estimation from multi-view geometry requires salient features, which is prone to error on the road or in challenging lighting conditions. In contrast, LiDAR sensors can readily capture the depth information of the objects, but as they are often equipped with few optical channels (e.g. 16), it can be difficult to infer the underlying geometry in real time due to their sparse resolution. To alleviate this problem, our method leverages centimeter-level localization \cite{fsr19:avl} to extract small dense regions of a previously built dense point cloud map. These smaller regions are then projected into the semantically segmented image to retrieve depth information. Building such a dense point map can be automated and only requires driving through the area once and is thus less expensive than human labeling.

With localization in place, the intrinsic camera parameters and the relative transformation between the camera and LiDAR are used to project point cloud data into the 2D image space. The point cloud data is then associated semantically using the nearest neighbor search. The relative transformation between the camera and the LiDAR is estimated using the PnP method \cite{epnp}, and the intrinsic matrix of the camera is determined by a traditional chessboard method. 

\subsection{Semantic Mapping}

While a point cloud with semantic labels naturally preserves the 3D geometry of the environment, such representation of the scene is subject to the sensor measurement noise and small semantic label fluctuations. To address this, we maintain a local or global probabilistic map, where the local map can provide direct dense semantic cues around the ego-vehicle and the global map can help automate the process of building HD maps. Semantic occupancy grids are used for both local and global semantic maps while the main difference is the reference frame. Our quantitative comparisons are performed in the global frame.

\begin{figure}
    \centering
    
    \begin{subfigure}
        \centering
        \includegraphics[scale=0.23]{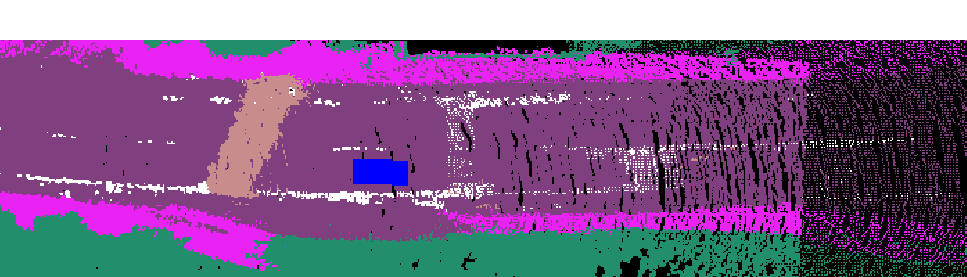}
    \end{subfigure}
    
    \begin{subfigure}
        \centering
        \includegraphics[scale=0.63]{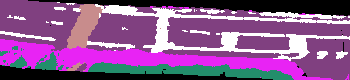}
    \end{subfigure}
    
    \caption{The map generation result of our algorithm. Top image is the local map and the blue car indicates the position of ego vehicles. Bottom image is the same region in global map.}
    \label{fig:local_and_global_map}
\end{figure}

A local probabilistic map is a bird's eye view representation in the body frame (rear-axle) of the ego vehicle. We build a local map for a given frame $i$, with the origin defined by the corresponding pose, and update it by using the semantic point cloud. Only when the difference of our new pose and old pose is beyond a threshold do we construct a new map and transform the previous map to account for vehicle movement. In contrast, a global probabilistic map operates directly in the global frame without the need of map transformations. A side-by-side visual comparison is shown in Figure \ref{fig:local_and_global_map}; where the top image corresponds to a local frame representation and the bottom image corresponds to a global frame representation. 

The semantic occupancy grid has height $H$, width $W$, and channels $C$. Each channel corresponds to a semantic class of the scene. When the semantic point cloud is constructed, we project it onto the grid using the $x$ and $y$ components. The semantic label of the point will be regarded as the observed semantic label of its nearest cell $c_{ij}$. Every cell in the grid covers a $d\times d$ square area (in meters) of the physical world where the value $d$ is a discretization factor. 

The robustness of the semantic occupancy grid estimation is enhanced by a probabilistic model which leverages both the semantic and LiDAR intensity information from the point cloud to reduce the prediction error. We denote the semantic label distribution across all the channels as $\mathbf{S}_t$, the observed semantic labels as $\mathbf{z}_t$, and observed LiDAR intensity as $\mathcal{I}_t$. Hence, the task is to estimate $\mathbf{S}_t$ from our past observations, i.e. the probability distribution of $P\left(\mathbf{S}_t|\mathbf{z}_{1:t}, \mathcal{I}_{1:t} \right)$. By following the Markov assumption and assuming that observed semantic labels and LiDAR intensity are conditionally independent given $\mathbf{S}_t$, the update rule of the semantic probability is 
\begin{align*}
    &{P}\left(\mathbf{S}_t|\mathbf{z}_{1:t}, \mathcal{I}_{1:t} \right) \\\hspace*{\fill} 
    =&\frac{1}{\mathbf{Z}}P\left(\mathbf{z}_t|\mathbf{S}_t \right)P\left(\mathcal{I}_t|\mathbf{S}_t \right)P\left(\mathbf{S}_{t-1}|\mathbf{z}_{1:t-1}, \mathcal{I}_{1:t-1} \right) 
\end{align*}

where $\mathbf{Z}$ is a normalization factor. Here we also assume that $P\left(\mathbf{S}_t|\mathbf{z}_{1:t-1}, \mathcal{I}_{1:t-1}\right)=P\left(\mathbf{S}_{t-1}|\mathbf{z}_{1:t-1}, \mathcal{I}_{1:t-1}\right)$. We model $P\left(\mathbf{z}_t|\mathbf{S}_t \right)$ with a 2D matrix $\mathbf{M}$ where an element in the $i$th row and $j$th column represents the likelihood of label $i$ being predicted as label $j$. It characterizes our confidence of prediction to allow a more accurate probabilistic update. We model $P\left(\mathcal{I}_t|\mathbf{S}_t \right)$ as a function of the reflectivity rate of each class in the scene.

The intensity from LiDAR sensors serves a strong cue for different materials on the scene. For example, the top image in Figure \ref{fig:pcd_pred_gt} illustrates a BEV intensity map on a road segment. Because lane marks are painted white, they can reflect light at higher intensity and thus be segmented out with a threshold $k$. We can use this as a prior to reason about the layout of the scene: it can help for the cases when semantic segmentation fails to capture the true label in poor lighting conditions.
\begin{table}[ht]
\centering
\begin{tabular}{||c c c c||} 
    \hline
    \multicolumn{4}{|c|}{Training Labels} \\
    \hline\hline
    \textcolor{c_curb}{curb} & \textcolor{c_crosswalk}{crosswalk} & \textcolor{c_road}{road} & \textcolor{c_sidewalk}{sidewalk} \\
    \textcolor{c_building}{building} & \textcolor{c_person}{person} & \textcolor{c_bicyclist}{bicyclist} & \textcolor{c_motorcyclist}{motorcyclist} \\
    lane marking (white) & \textcolor{c_sky}{sky} & \textcolor{c_vegetation}{vegetation} & \textcolor{c_manhole}{manhole} \\
    \textcolor{c_pole}{pole} & \textcolor{c_traffic_sign}{traffic-sign} & \textcolor{c_bicycle}{bicycle} & \textcolor{c_bus}{bus} \\
    \textcolor{c_car}{car} & \textcolor{c_motorcycle}{motorcycle} & \textcolor{c_truck}{truck} & \\
    \hline
\end{tabular}
\caption{Training Labels and Their Associated Colors for the Semantic Segmentation Network}
\label{table:training_labels}
\end{table}

\section{EXPERIMENTS}

\begin{table*}[]
    \centering
    \begin{tabular}{c|ccc|c|ccc}
        \multicolumn{8}{c}{ } \\
         \multirow{2}{*}{Methods}&\multicolumn{3}{c|}{IoU}&\multirow{2}{*}{mIoU}&\multicolumn{3}{c}{Accuracy}  \\
         &roads&crosswalks&lane marks&&roads&crosswalks&lane marks\\
         \hline 
         \hline
         Vanilla&\textbf{0.715}&0.537&0.135&0.462&\textbf{0.797}&0.577&0.180\\ \hline
         Vanilla+I&0.712&0.510&0.163&0.462&0.789&0.548&0.234\\ \hline
         CFN&0.671&\textbf{0.605}&\textbf{0.301}&\textbf{0.526}&0.708&\textbf{0.696}&0.652\\ \hline
         CFN+I&0.669&0.588&0.298&0.518&0.705&0.676&\textbf{0.657}\\ \hline
    \end{tabular}
    \caption{Quantitative evaluation on our labeled data for road, crosswalk and lane mark regions. Refer to section \ref{section_semantic_mapping} for details.}
    \label{table:experiment_results}
\end{table*}

\begin{figure}[h]
    \centering
    \includegraphics[scale=0.15]{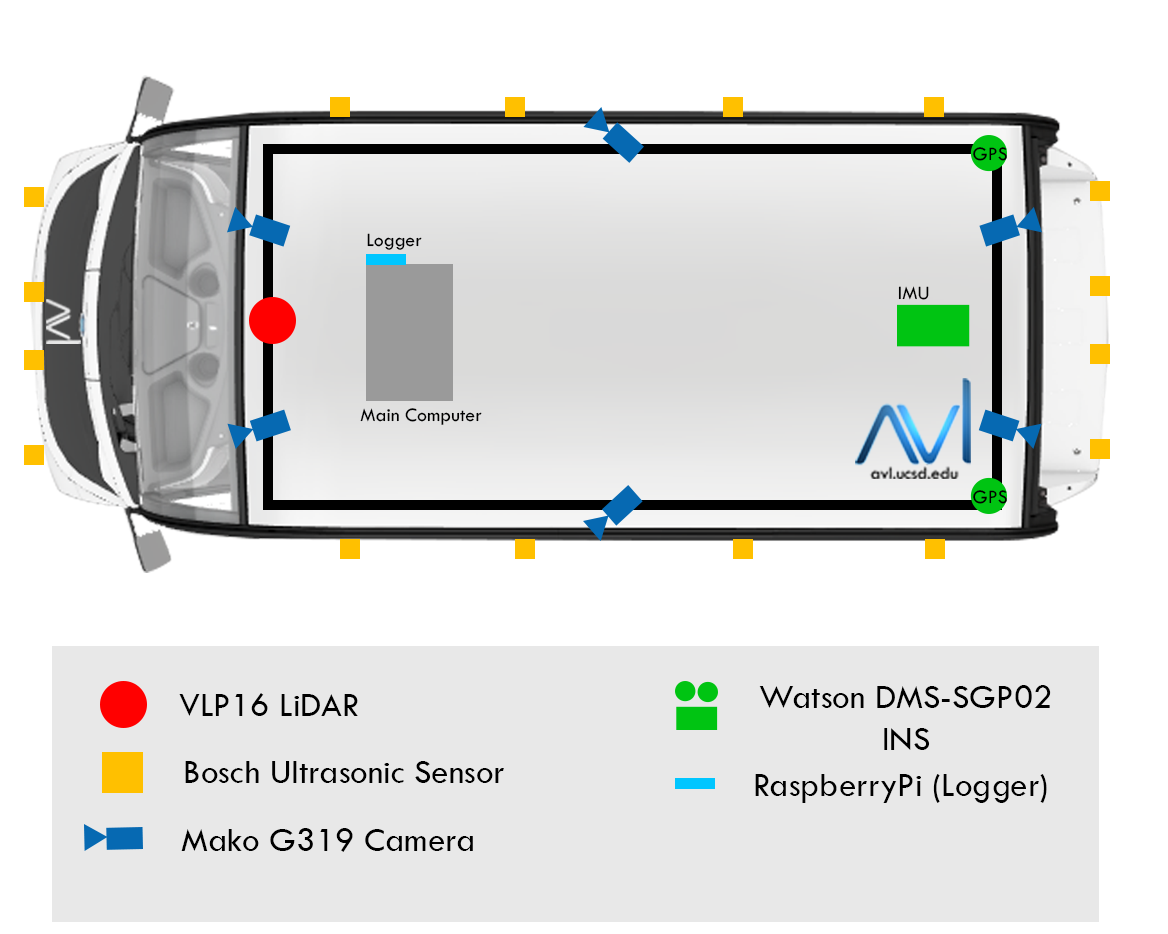}
    \caption{Vehicle Sensor Configuration}
    \label{fig:vehicle layout}
\end{figure}

Our experimental data was collected by one of our experimental autonomous cars \cite{fsr19:avl}. The car is equipped with a 16-channel LiDAR and six cameras. The cameras are set up as two on the front, one on each side and two on the back as shown in Figure \ref{fig:vehicle layout}. Data from the front left camera, LiDAR, and vehicle position is recorded for experiments by driving along multiple areas at the UC San Diego campus. The camera data is streamed at approximately 13 Hz and the LiDAR scans at approximately 10 Hz. We drive through the campus to collect data for urban driving scenarios--including challenging scenarios like driving along steep hills, intersections and construction sites.

\subsection{Image Semantic segmentation}
\label{sec:exp_semantic}

\subsubsection{Training Dataset}
We have 18,000 training images and 2,000 validation images from the Mapillary data set. We merge terrain into vegetation, different types of riders into the human category, traffic-sign-back and traffic-sign-front into traffic-sign, bridge into building, and different kinds of crosswalks into a single crosswalk class. The training dataset is augmented by random horizontal flips with 0.5 probability, random resize with the scale ranging from 0.5 to 2, and random crop. These images are also normalized to a distribution with mean of (0.485, 0.456, 0.406) and standard deviation of (0.229, 0.224, 0.225). Due to the similarity of the Mapillary dataset and our driving scenarios, as well as the intense data augmentation in the training process, we do not visually observe severe performance drop when testing it in the UC San Diego campus.

\subsubsection{Hyperparameters}
We use a batch size of 16 with synchronized batch normalization \cite{Zhao_2017_PSPNet} to train our network for 200 epochs on eight 2080Ti GPUs with input image sizes of 640x640. The output stride of the network is eight. We use a SGD optimizer and employ a polynomial learning rate policy \cite{Chen_2018_DeeplabV3Plus, Zhu_2019_nvidia_label_relaxation} where the learning rate is $base\_lr \times (1-\frac{epoch}{max epoch})^{power}$ with 0.005 base learning rate and power=0.9. The momentum and weight decay are set to 0.9 and $4e^{-5}$, respectively.

\subsubsection{Metric}
mIoU is used to evaluate the performance of the network. The mIoU of ResNeXt50 in the validation set is 68.32\%. Compared with ResNet101, ResNeXt50's performance slightly decreases but requires much less memory (from 367MB to 210MB), which is preferable for our onboard hardware where the memory is limited. The inference time of the network is approximately 0.2s per image.

\subsection{Semantic Mapping}\label{section_semantic_mapping}

We select a region that spans $1.1$ km of the UC San Diego campus to evaluate our map generation results. An HD map has been manually annotated in this area. It contains road information such as crosswalks, sidewalks, and center of road lane definitions and has been tested in realistic environments. Based on these labels, we generate our semantic map with five channels: \textit{road, crosswalk, lane marks, vegetation, and sidewalk} with resolution $d=0.2$ meters. The labeling of HD map is not a trivial work. However, it exactly showcases the value of being able to automate the entire process. 

The mIoU and pixel accuracy are used as the metrics of our evaluation. While mIoU reflects our model's segmentation recall and precision, it is worth noting these metrics are influenced by the sparsity of the LiDAR point cloud: the model output may be very accurate but may contain a lot of unclassified cells (holes). We address this problem by using a smoothing kernel to interpolate the missing labels on our map.

For all the experiments in Table \ref{table:experiment_results}, we clip the local dense point maps extracted up to $15$m along the longitudinal axis and $-10$ to $10$ m along the lateral axis of the vehicle as the semantic segmentation performance drops significantly for a longer range.

\subsubsection{Modeling of Observation Uncertainty}

We first validate the design of matrix $\mathbf{M}$ that models the observation uncertainty of the semantic label $\mathbf{z}_t$ given $\mathbf{S}_t$. For numerical stability reasons, we represent the elements of $\mathbf{M}$ logarithmically. We investigate two approaches: one is defined by $\mu\left(\mathbf{I}+\lambda \mathbf{1}\right)$ where $\lambda$ is a hyper-parameter and $\mu$ is a normalization factor: we refer to this model as Vanilla. Another approach is the confusion matrix of the semantic segmentation network in the Mapillary validation data set: we define this as CFN. During the inference time, we assign each cell to the label with the highest probability. Quantitative results are shown in Table \ref{table:experiment_results}. Compared with Vanilla version, CFN has a significant improvement in IoU and pixel accuracy on the classes of crosswalks and lane marks. This indicates the advantage of utilizing the confusion matrix of the network to model the prediction error in the semantic segmentation and thus provides a better map generation result. 

\subsubsection{Integration with LiDAR Intensity}

\begin{figure}[ht]
    \centering
    \includegraphics[scale=1.1]{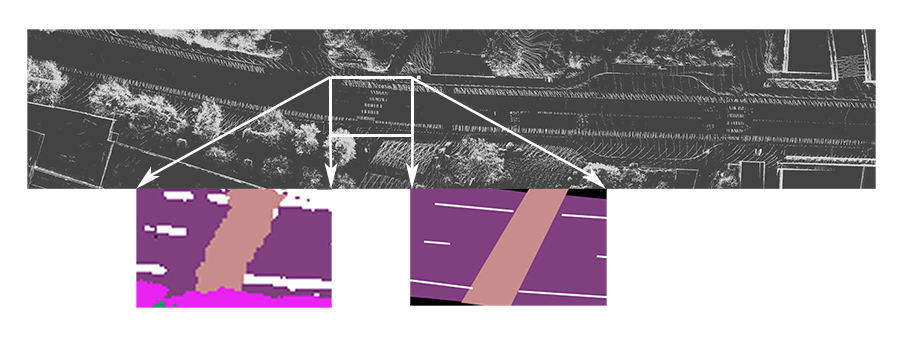}
    \caption{A visualization of our generated map (bottom left), the ground truth label (bottom right), and the LiDAR point cloud map (top). The point cloud map has been thresholded based on their intensity value. Any points that have intensity below a certain threshold $k$ are discarded. With this threshold, we can clearly see the layout of the crosswalk and lane marks on the road.}
    \label{fig:pcd_pred_gt}
\end{figure}

In order to utilize the fact that different materials on the road have different reflectivity, we first remove all the intensity data that is less than $k=14$ (e.g. Figure \ref{fig:pcd_pred_gt}); this normalized threshold was manually calibrated for a Velodyne VLP-16 LiDAR. Then during the semantic mapping step, if a label is predicted as lane marks, we increase its logarithmic probability by a constant factor $\gamma$. This essentially suppresses our prediction of other classes and increases our prediction confidence for lane marks. The models that contain the integration of intensity have a "+I" in Table \ref{table:experiment_results}. For Vanilla+I, we observe better results with respect to Vanilla in terms of accuracy and IoU of the lane marks but slightly decreases for roads and crosswalks, indicating the benefit of intensity integration for lane mark prediction. This tendency, however, does not replicate for CFN+I with respect to CFN. To achieve further improvement, a more sophisticated function may be needed to model the LiDAR intensity. 

\begin{figure}[ht]
    \centering
    \includegraphics[scale=1.0]{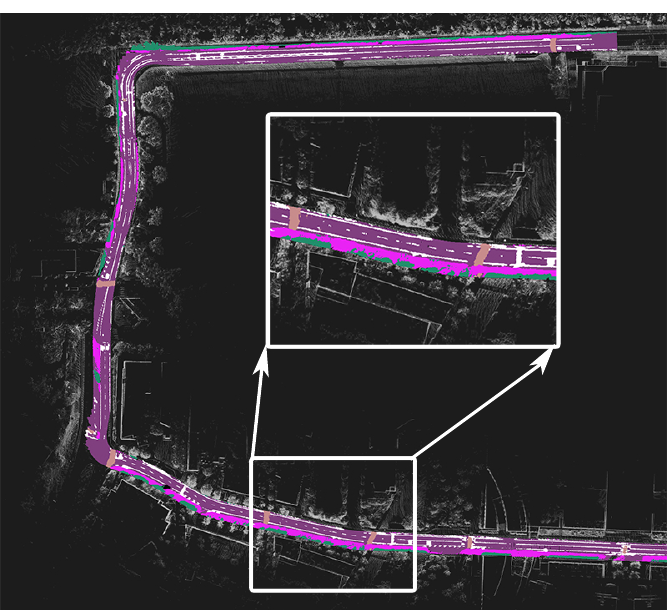}
    \caption{The BEV version of our generated map in the entire region of our testing data set. Our semantic map is displayed on top of the dense point cloud map. One segment of the map has been magnified to show the details of the map. Best view in color.}
    \label{fig:stitched_with_pcd}
\end{figure}

An example of the global map generated by our CFN+I model for the entire test region is shown in Figure \ref{fig:stitched_with_pcd}. A region of the map has been amplified in the figure, showing that our model can capture the static elements on the road clearly. 

\subsection{Comparison to Sparse LiDAR Scan}

\begin{figure}
    \centering
    \begin{subfigure}
        \centering
        \includegraphics[scale=0.23]{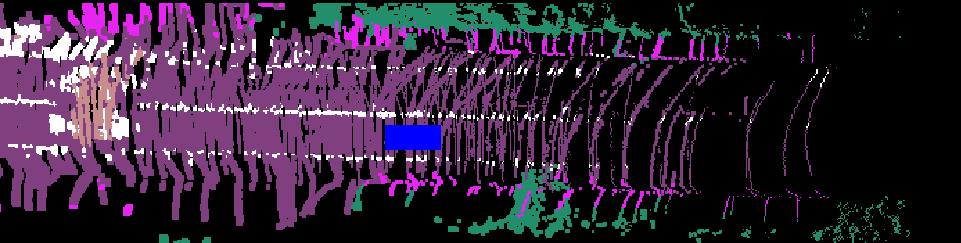}
    \end{subfigure}
    \caption{The semantic map generated from real time LiDAR scan. The black region on the scene indicates unknown area which is not detected by the LiDAR sensor.}
    \label{fig:points_raw_sparse}
\end{figure}

One possible alternative for associating semantic images with depth information is to use the point cloud data generated by the LiDAR in real time. By following a similar mapping approach, we project the point cloud onto the semantic image frame and build the semantic map. Figure \ref{fig:points_raw_sparse} shows that this approach gives real-time performance. However, for the 16-channel LiDAR used, the point cloud scans are too sparse to construct a semantic map at greater distances. This becomes worse when the car drives faster. Therefore, a pre-built dense point cloud map allows us to construct semantic maps for longer ranges.

\subsection{Comparison to Planar Assumption}

\begin{figure}
    \centering
    \begin{subfigure}
        \centering
        \includegraphics[scale=0.23]{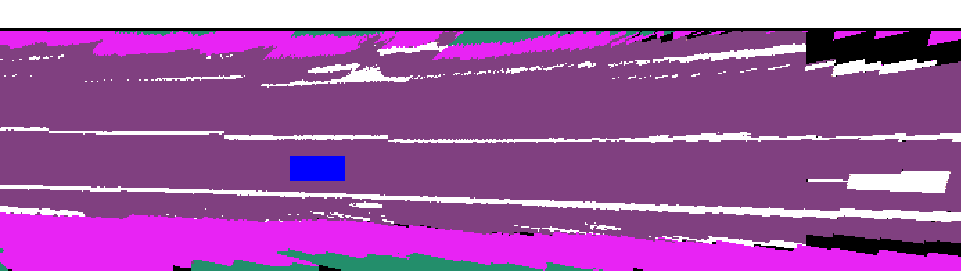}
    \end{subfigure}
    \caption{The semantic map generated by back-projecting 2D semantic image into the 3D space with planar assumption. The map presented in the vehicle frame. The distortion of lines indicates the failure case.}
    \label{fig:planar assumption}
\end{figure}

Another method we explored is back-projecting the 2D semantic image into the 3D space using a homography with the assumption that the ground is flat. The back-projection approach leaves no black holes. Nevertheless, the planar assumption fails along steep hills or road intersections in urban driving scenarios as shown in Figure \ref{fig:planar assumption}: this leads to considerable distortion at longer ranges.

\section{CONCLUSION}
By fusing the rich information from semantic labels on image frames, our comparisons to manually annotated maps indicate that this work effectively introduces a statistical method for identifying road features and localizing them in bird's eye view. This method can be extended for automating HD map annotation for crosswalks, lane markings, drivable surfaces, and sidewalks. These features can be incorporated for generating HD maps independently of predefined HD map formats with the additional extension of center lane identifications which are often used for path tracking algorithms. 

By accounting for the road network junctions and forks, future work involves the full automation of road network annotations that could leverage graphical methods. While a combination of the techniques proposed can potentially address the scalability drawbacks from HD maps, they also propose new areas of research on high-level dynamic planning. Currently, many autonomous driving architectures require dense point cloud maps for localization and come at a scalability and maintenance cost in a similar way that HD maps do. By dynamically estimating drivable surfaces, traffic lanes, lane markings and other road features, the notion of using centimeter-level localization could be removed as long as immediate actions can be extracted from a high-level planner. In future work, we plan to seek out solutions for fully automating the HD mapping process while exploring the idea of dynamic planning without a detailed dense point cloud map. 

\bibliographystyle{unsrt}
\bibliography{citation}

\begin{thebibliography}{10}

\bibitem{Jiao18MLHDMap}
J.~{Jiao}.
\newblock Machine learning assisted high-definition map creation.
\newblock In {\em 2018 IEEE 42nd Annual Computer Software and Applications
  Conference (COMPSAC)}, volume~01, pages 367--373, July 2018.

\bibitem{Douillard11Classification}
B.~Douillard, D.~Fox, F.~Ramos, and H.~Durrant-Whyte.
\newblock Classification and semantic mapping of urban environments.
\newblock {\em The International Journal of Robotics Research}, 30(1):5--32,
  2011.

\bibitem{Sengupta12DenseVisual}
S.~{Sengupta}, P.~{Sturgess}, L.~{Ladický}, and P.~H.~S. {Torr}.
\newblock Automatic dense visual semantic mapping from street-level imagery.
\newblock In {\em 2012 IEEE/RSJ International Conference on Intelligent Robots
  and Systems}, pages 857--862, Oct 2012.

\bibitem{Long_2015_CVPR_FCNNet}
Jonathan Long, Evan Shelhamer, and Trevor Darrell.
\newblock Fully convolutional networks for semantic segmentation.
\newblock In {\em The IEEE Conference on Computer Vision and Pattern
  Recognition (CVPR)}, June 2015.

\bibitem{Zhao_2017_PSPNet}
Hengshuang Zhao, Jianping Shi, Xiaojuan Qi, Xiaogang Wang, and Jiaya Jia.
\newblock Pyramid scene parsing network.
\newblock {\em 2017 IEEE Conference on Computer Vision and Pattern Recognition
  (CVPR)}, Jul 2017.

\bibitem{Chen_2018_DeeplabV3Plus}
Liang-Chieh Chen, Yukun Zhu, George Papandreou, Florian Schroff, and Hartwig
  Adam.
\newblock Encoder-decoder with atrous separable convolution for semantic image
  segmentation.
\newblock {\em Lecture Notes in Computer Science}, page 833–851, 2018.

\bibitem{Maturana18RealtimeSemantic}
Daniel Maturana, Po-Wei Chou, Masashi Uenoyama, and Sebastian Scherer.
\newblock Real-time semantic mapping for autonomous off-road navigation.
\newblock In Marco Hutter and Roland Siegwart, editors, {\em Field and Service
  Robotics}, pages 335--350, Cham, 2018. Springer International Publishing.

\bibitem{mattyus2017deeproadmapper}
Gell{\'e}rt M{\'a}ttyus, Wenjie Luo, and Raquel Urtasun.
\newblock Deeproadmapper: Extracting road topology from aerial images.
\newblock In {\em Proceedings of the IEEE International Conference on Computer
  Vision}, pages 3438--3446, 2017.

\bibitem{homayounfar2019dagmapper}
Namdar Homayounfar, Wei-Chiu Ma, Justin Liang, Xinyu Wu, Jack Fan, and Raquel
  Urtasun.
\newblock Dagmapper: Learning to map by discovering lane topology.
\newblock In {\em Proceedings of the IEEE International Conference on Computer
  Vision}, pages 2911--2920, 2019.

\bibitem{Cordts_2016_CVPR_cityscapes}
Marius Cordts, Mohamed Omran, Sebastian Ramos, Timo Rehfeld, Markus Enzweiler,
  Rodrigo Benenson, Uwe Franke, Stefan Roth, and Bernt Schiele.
\newblock The cityscapes dataset for semantic urban scene understanding.
\newblock In {\em The IEEE Conference on Computer Vision and Pattern
  Recognition (CVPR)}, June 2016.

\bibitem{BrostowFC:PRL2008_Camvid}
Gabriel~J. Brostow, Julien Fauqueur, and Roberto Cipolla.
\newblock Semantic object classes in video: A high-definition ground truth
  database.
\newblock {\em Pattern Recognition Letters}, xx(x):xx--xx, 2008.

\bibitem{MVD2017_Mapillary_Vistas}
Gerhard Neuhold, Tobias Ollmann, Samuel Rota~Bul\`o, and Peter Kontschieder.
\newblock The mapillary vistas dataset for semantic understanding of street
  scenes.
\newblock In {\em International Conference on Computer Vision (ICCV)}, 2017.

\bibitem{chen2017rethinking_deeplabv3}
Liang-Chieh Chen, George Papandreou, Florian Schroff, and Hartwig Adam.
\newblock Rethinking atrous convolution for semantic image segmentation, 2017.

\bibitem{Wu_2018_squeezeseg}
Bichen Wu, Alvin Wan, Xiangyu Yue, and Kurt Keutzer.
\newblock Squeezeseg: Convolutional neural nets with recurrent crf for
  real-time road-object segmentation from 3d lidar point cloud.
\newblock {\em 2018 IEEE International Conference on Robotics and Automation
  (ICRA)}, May 2018.

\bibitem{heinzler2020cnn_rangenet}
Robin Heinzler, Florian Piewak, Philipp Schindler, and Wilhelm Stork.
\newblock Cnn-based lidar point cloud de-noising in adverse weather.
\newblock {\em IEEE Robotics and Automation Letters}, 5(2):2514--2521, 2020.

\bibitem{wang2018pointseg}
Yuan Wang, Tianyue Shi, Peng Yun, Lei Tai, and Ming Liu.
\newblock Pointseg: Real-time semantic segmentation based on 3d lidar point
  cloud, 2018.

\bibitem{Tchapmi_2017_segcloud}
Lyne Tchapmi, Christopher Choy, Iro Armeni, JunYoung Gwak, and Silvio Savarese.
\newblock Segcloud: Semantic segmentation of 3d point clouds.
\newblock {\em 2017 International Conference on 3D Vision (3DV)}, Oct 2017.

\bibitem{KOSTAVELIS201586}
Ioannis Kostavelis and Antonios Gasteratos.
\newblock Semantic mapping for mobile robotics tasks: A survey.
\newblock {\em Robotics and Autonomous Systems}, 66:86 -- 103, 2015.

\bibitem{Wolf08SemanticMapping}
D.~F. {Wolf} and G.~S. {Sukhatme}.
\newblock Semantic mapping using mobile robots.
\newblock {\em IEEE Transactions on Robotics}, 24(2):245--258, April 2008.

\bibitem{Sengupta13Stereo}
S.~{Sengupta}, E.~{Greveson}, A.~{Shahrokni}, and P.~H.~S. {Torr}.
\newblock Urban 3d semantic modelling using stereo vision.
\newblock In {\em 2013 IEEE International Conference on Robotics and
  Automation}, pages 580--585, May 2013.

\bibitem{Levinson10ProbMap}
J.~{Levinson} and S.~{Thrun}.
\newblock Robust vehicle localization in urban environments using probabilistic
  maps.
\newblock In {\em 2010 IEEE International Conference on Robotics and
  Automation}, pages 4372--4378, May 2010.

\bibitem{Shao2017HIGHACCURACYVL}
Yunming Shao, Charles Toth, Dorota~A. Grejner-Brzezinska, and Lowber~B.
  Strange.
\newblock High-accuracy vehicle localization using a pre-built probability map.
\newblock 2017.

\bibitem{Wu18ProbMapPedes}
J.~{Wu}, J.~{Ruenz}, and M.~{Althoff}.
\newblock Probabilistic map-based pedestrian motion prediction taking traffic
  participants into consideration.
\newblock In {\em 2018 IEEE Intelligent Vehicles Symposium (IV)}, pages
  1285--1292, June 2018.

\bibitem{Xie_2017_ResNeXt}
Saining Xie, Ross Girshick, Piotr Dollar, Zhuowen Tu, and Kaiming He.
\newblock Aggregated residual transformations for deep neural networks.
\newblock {\em 2017 IEEE Conference on Computer Vision and Pattern Recognition
  (CVPR)}, Jul 2017.

\bibitem{ImageNet2015}
Olga Russakovsky, Jia Deng, Hao Su, Jonathan Krause, Sanjeev Satheesh, Sean Ma,
  Zhiheng Huang, Andrej Karpathy, Aditya Khosla, Michael Bernstein,
  Alexander~C. Berg, and Li~Fei-Fei.
\newblock {ImageNet Large Scale Visual Recognition Challenge}.
\newblock {\em International Journal of Computer Vision (IJCV)},
  115(3):211--252, 2015.

\bibitem{He_2016_resnet}
Kaiming He, Xiangyu Zhang, Shaoqing Ren, and Jian Sun.
\newblock Deep residual learning for image recognition.
\newblock {\em 2016 IEEE Conference on Computer Vision and Pattern Recognition
  (CVPR)}, Jun 2016.

\bibitem{Chollet_2017_Xception}
Francois Chollet.
\newblock Xception: Deep learning with depthwise separable convolutions.
\newblock {\em 2017 IEEE Conference on Computer Vision and Pattern Recognition
  (CVPR)}, Jul 2017.

\bibitem{fsr19:avl}
David Paz, Po-Jung Lai, Sumukha Harish, Hengyuan Zhang, Nathan Chan, Chun Hu,
  Sumit Binnani, and Henrik Christensen.
\newblock Lessons learned from deploying autonomous vehicles at {UC San Diego}.
\newblock In {\em Field and Service Robotics}, Tokyo, JP, August 2019.

\bibitem{epnp}
Vincent Lepetit, Francesc Moreno-Noguer, and Pascal Fua.
\newblock Epnp: An accurate o(n) solution to the pnp problem.
\newblock {\em International Journal of Computer Vision}, 81, 02 2009.

\bibitem{Zhu_2019_nvidia_label_relaxation}
Yi~Zhu, Karan Sapra, Fitsum~A. Reda, Kevin~J. Shih, Shawn Newsam, Andrew Tao,
  and Bryan Catanzaro.
\newblock Improving semantic segmentation via video propagation and label
  relaxation.
\newblock {\em 2019 IEEE/CVF Conference on Computer Vision and Pattern
  Recognition (CVPR)}, Jun 2019.

\end{thebibliography}

\end{document}